\documentclass[twoside,11pt]{article}

\usepackage{blindtext}

\usepackage{jmlr2e}

\usepackage{xcolor}

\usepackage{booktabs}
\usepackage{multirow}
  
\usepackage{makecell} 
\usepackage{diagbox}

\usepackage{pifont}
\newcommand{\cmark}{\ding{51}}
\newcommand{\xmark}{\ding{55}}

\newcommand{\Name}{{Fast-PGM}}

\usepackage{lastpage}
\ShortHeadings{\Name{}: Fast Probabilistic Graphical Model Learning and Inference}{Jiang, Wen, Yang, Mansoor and Mian}
\firstpageno{1}

\begin{document}

\title{\Name{}: Fast Probabilistic Graphical Model Learning and Inference}

% \author{\name Author One \email one@stat.washington.edu \\
%        \addr Department of Statistics\\
%        University of Washington\\
%        Seattle, WA 98195-4322, USA
%        \AND
%        \name Author Two \email two@cs.berkeley.edu \\
%        \addr Division of Computer Science\\
%        University of California\\
%        Berkeley, CA 94720-1776, USA}

\author{\name Jiantong Jiang$^{1}$ \email jiantong.jiang@research.uwa.edu.au \\
       \name Zeyi Wen$^{2,3}$ \email wenzeyi@ust.hk \\
       \name Peiyu Yang$^{1}$ \email peiyu.yang@research.uwa.edu.au \\
       \name Atif Mansoor$^{1}$ \email atif.masoor@uwa.edu.au \\
       \name Ajmal Mian$^{1}$ \email ajmal.mian@uwa.edu.au \\
       \addr $^{1}$Department of CSSE, The University of Western Australia, Australia \\
       $^{2}$Hong Kong University of Science and Technology (Guangzhou), $^{3}$HKUST, China}

\editor{TBD}

\maketitle

\begin{abstract}%   <- trailing '%' for backward compatibility of .sty file
Probabilistic graphical models (PGMs) serve as a powerful framework for modeling complex systems with uncertainty and extracting valuable insights from data. However, users face challenges when applying PGMs to their problems in terms of efficiency and usability. This paper presents \textit{\Name{}}, an efficient and open-source library for PGM learning and inference. \Name{} supports comprehensive tasks on PGMs, including structure and parameter learning, as well as exact and approximate inference, and enhances efficiency of the tasks through computational and memory optimizations and parallelization techniques. Concurrently, \Name{} furnishes developers with flexible building blocks, furnishes learners with detailed documentation, and affords non-experts user-friendly interfaces, thereby ameliorating the usability of PGMs to users across a spectrum of expertise levels. 
%Experimental results demonstrate the efficiency of \Name{} over existing systems. 
The source code of \Name{} is available at \url{https://github.com/jjiantong/FastPGM}.

\end{abstract}


\section{Introduction}

Probabilistic graphical models (PGMs)~\citep{pgm} employ a transparent and intuitive representation to compactly encode random variables and their interactions, while offering a mathematically sound framework grounded in probability theory~\citep{prob}. PGMs have found widespread application across domains such as biomedical informatics~\citep{bnapp2,bnapp3,mnapp5}, computer vision~\citep{mnapp2, mnapp3, mnapp4}, environmental monitoring~\citep{bnapp5,bnapp6}, risk management~\citep{bnapp1, bnapp4} and transportation~\citep{bnapp7, mnapp1}.
However, realizing the full potential of PGMs in practical scenarios requires addressing notable challenges related to efficiency and usability. On the efficiency front, both learning and inference processes in PGMs are expensive due to their high computational complexity~\citep{chickering2004large, bni, Dagum1993} and the irregular nature of the graphical structures, thereby impeding their application in large or complex problems. As for the usability challenges, many existing libraries are constrained by their limited support of various PGM tasks and algorithms. Additionally, the support for complete open-source availability, user-friendly interfaces, and comprehensive documentation is demanding, especially for domain experts who may lack sufficient expertise in machine learning techniques.

This paper proposes \Name{} as a solution to address the challenges above. \Name{} has the following crucial features:
(i) \textbf{Broad Task Support.} \Name{} supports all the key tasks on PGMs, encompassing structure and parameter learning, as well as exact and approximate inference, capable of handling various problems with diverse requirements.
(ii) \textbf{High Efficiency.} We exploit different parallelization techniques tailored to different tasks to improve the efficiency of \Name{}. Meanwhile, we develop memory management and computation simplification optimizations to enhance data locality and alleviate the computational burden.
(iii) \textbf{Ease of Use.} \Name{} offers a modular design with various functionalities to allow quick and easy customization across all modules. \Name{} also provides user-friendly interfaces and documentation to ensure accessibility. The Python interfaces further facilitate integration within the enormous Python ecosystem, including seamless compatibility with tasks like hyperparameter optimization~\citep{synetune}.

% \textbf{F1: Broad Task Support.} \Name{} supports all the key tasks on PGMs, encompassing structure learning, parameter learning, exact inference, and approximate inference. This enables \Name{} to handle a wide range of problems, accommodating diverse requirements and complexities.
% %
% \textbf{F2: High Efficiency.} We exploit various parallelization techniques on multi-core CPUs tailored to different tasks to improve the efficiency of \Name{}. Meanwhile, we develop memory management optimizations and computation simplification strategies to enhance data locality and alleviate the computational burden.
% %
% \textbf{F3: Ease of Use.} \Name{} offers various functionalities and interfaces to allow quick and easy optimization, extension, and customization across all PGM learning and inference modules. Additionally, \Name{} provides user-friendly interfaces, documentation, and visualization, ensuring accessibility for of varying expertise levels. The Python interfaces further facilitate integration within the enormous Python ecosystem, including seamless compatibility with tasks such as hyperparameter optimization.

\begin{figure}%[h]
	\centering
	\includegraphics[width=0.8\textwidth]{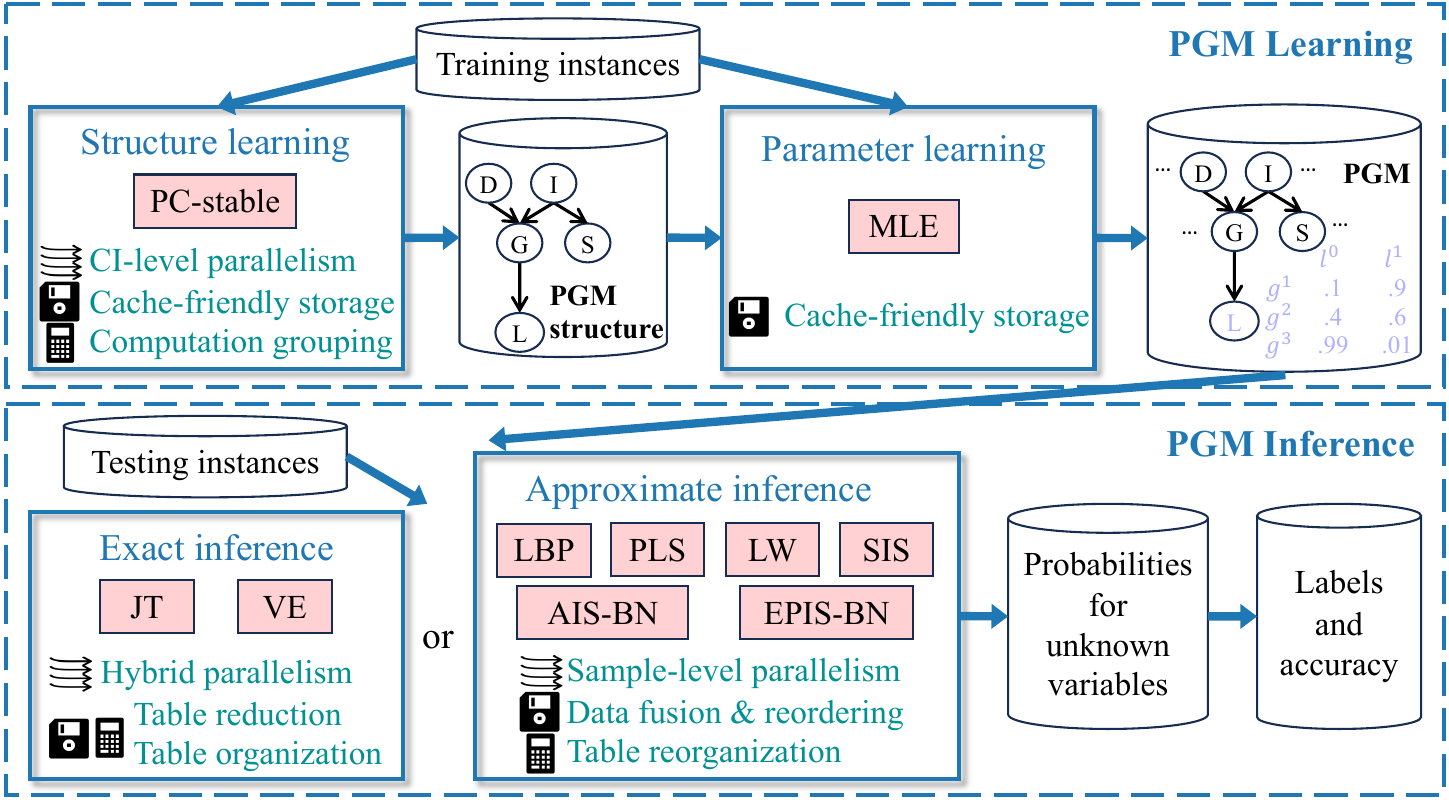}
	\caption{The overview of PGM learning and inference in \Name{}.}
	\label{fig_overview}
\end{figure}

\section{Functionality Design Overview}

Figure~\ref{fig_overview} shows the overview of \Name{} with supported tasks, algorithms, and optimizations.
\Name{} supports all the fundamental tasks associated with PGMs. For \textit{structure learning}, \Name{} implements the renowned PC-stable algorithm~\citep{pc, pc2} to find the underlying graph that best matches the training instances. Following structure acquisition, \textit{parameter learning} is performed to quantify the probabilistic relationships encoded by the structure through maximum likelihood estimation. The above learning process yields a refined PGM that matches the training instances. Subsequently, PGM \textit{inference} aims to calculate the posterior distribution of all the unknown variables, which is categorized into \textit{exact} and \textit{approximate} methods. For exact inference, \Name{} implements junction tree~\citep{jt} and variable elimination~\citep{Zhang1994}. Alternatively, for approximate inference, \Name{} supports an array of methods including loopy belief propagation~\citep{lbp}, probabilistic logic sampling~\citep{pls}, likelihood weighting~\citep{lw}, self-importance sampling~\citep{sis, sisv1}, adaptive importance sampling (AIS-BN)~\citep{aisbn}, and evidence pre-propagation importance sampling (EPIS-BN)~\citep{episbn, episbn2}. The integration of these key tasks also results in a complete process of classification.

Beyond the primary functionalities, \Name{} extends its utility with a range of auxiliary features vital for the utilization of PGMs. For instance, \Name{} offers tools for generating sample sets from a PGM, facilitating format transformation across network representations, and provides metrics like structural Hamming distance~\citep{shd} for learning and Hellinger's distance for inference~\citep{h-distance}, streamlining the evaluation process.

\section{Technical Details}

To ensure high efficiency and modular design, \Name{} is implemented and accelerated in C/C++. On the one hand, \Name{} provides well-designed modular building blocks, various functionalities, and rich interfaces, enabling developers to quickly and easily optimize, extend, and customize across all PGM learning and inference modules. On the other hand, \Name{} offers user-friendly interfaces, documentation, and visualization, ensuring accessibility of the PGM system for learners and non-specialists. 
Furthermore, in line with the growing preference for Python in the machine learning community, \Name{} offers Python access to every learning and inference task on PGMs. The Python APIs facilitate integration within the extensive Python ecosystem. For example, \Name{} can be compatible with hyperparameter optimization techniques~\citep{synetune, fastbo} to provide automatic decisions on the proper algorithms and hyperparameters for different scenarios. 

Regarding efficiency, we adopt and extend the novel optimization techniques in our previous work~\citep{fast-bn, fast-bni, jiang2024fast} to accelerate \Name{}, as illustrated in Figure~\ref{fig_overview}. The key optimizations are outlined as follows: 
(i) We explore parallelism at the conditional independence level for structure learning to address the efficiency issues in current parallelization schemes. A dynamic work pool is also introduced to monitor processing and schedule workloads. 
(ii) We employ a cache-friendly data storage scheme for structure and parameter learning to improve memory efficiency.
(ii) We identify and group the similar and dependent computations within structure learning to reduce redundant operations and optimize computational resources.
(iv) We implement a hybrid inter- and intra-clique parallelism for exact inference. This involves a tree traversal method and a root selection strategy to maximize parallelization opportunities.
(v) We carefully reorganize the potential tables, a crucial underlying data structure in PGMs, before inference to optimize memory storage and reduce computational complexity.
(vi) We implement sample-level parallelism for approximate inference to improve overall efficiency.
(vii) We design the data fusion and data reordering techniques for approximate inference to enhance data locality. These techniques mitigate irregular memory accesses caused by the stochastic nature of sampling and the graphical structure of PGMs.

\section{Comparison to Existing Libraries}

% \begin{table*}
%     \centering
%     \caption{Feature comparison of various PGM-related libraries.}
%     \setlength{\tabcolsep}{0.95mm}{
%     \begin{tabular}{ccccccccc}
%         \toprule
%         Library & \textbf{ours} & SMILE & causal-learn & Tetrad & bnlearn & pcalg & ParallelPC & UnBBayes \\
%         \midrule
%         Language & C++ & C++ & Python & Java & R & R & R & Java \\
%         Open-source & \cmark & \xmark & \cmark & \cmark & \cmark & \cmark & \cmark & \cmark \\
%         Parallelization & \cmark & \xmark & \xmark & \xmark & \cmark & \xmark & \cmark & \xmark \\
%         Struct. learn. & \cmark & \cmark & \cmark & \cmark & \cmark & \cmark & \xmark & \cmark \\
%         Param. learn. & \cmark & \cmark & \xmark & \cmark & \cmark & \xmark & \xmark & \cmark \\
%         Ex. inf. & \cmark & \cmark & \xmark & \xmark & \cmark & \xmark & \xmark & \xmark \\
%         Appr. inf. & \cmark & \cmark & \xmark & \xmark & \cmark & \xmark & \xmark & \cmark \\
%         \bottomrule
%     \end{tabular}
%     }
%     \label{tab_libs}
% \end{table*}

\begin{table*}
    \centering
    \caption{Feature comparison of various PGM-related libraries.}
    \setlength{\tabcolsep}{1.8mm}{
    \begin{tabular}{ccccccccc}
        \toprule
        Library  & \makecell[c]{Structure\\learn.} & \makecell[c]{Param.\\learn.} & \makecell[c]{Ex.\\inf.} & \makecell[c]{Appr.\\inf.} & \makecell[c]{Open-\\source} & Parallel. & Update & Language\\
        \midrule
        causal-learn & \cmark & \xmark & \xmark & \xmark & \cmark & \xmark & \cmark & Python \\
        Tetrad & \cmark & \cmark & \xmark & \xmark & \cmark & \xmark & \cmark & Java \\
        ParallelPC & \cmark & \xmark & \xmark & \xmark & \cmark & \cmark & \xmark & R \\
        bnlearn & \cmark & \cmark & \cmark & \cmark & \cmark & \cmark & \cmark & R \\
        pcalg & \cmark & \xmark & \xmark & \xmark & \cmark & \xmark & \cmark & R \\
        FastInf & \xmark & \cmark & \cmark & \cmark & \xmark & \xmark & \xmark & C/C++ \\
        libDAI & \xmark & \cmark & \cmark & \cmark & \cmark & \xmark & \xmark & C/C++ \\
        UnBBayes & \cmark & \cmark & \xmark & \cmark & \cmark & \xmark & \cmark & Java \\
        Factorie & \xmark & \cmark & \xmark & \cmark & \cmark & \xmark & \xmark & Scala \\
        BNJ & \cmark & \xmark & \cmark & \cmark & \cmark & \xmark & \xmark & Java \\
        BNT & \cmark & \cmark & \cmark & \cmark & \cmark & \xmark & \xmark & Matlab \\
        SMILE & \cmark & \cmark & \cmark & \cmark & \xmark & \xmark & \cmark & C/C++ \\
        PNL & \cmark & \cmark & \cmark & \cmark & \cmark & \xmark & \xmark & C/C++ \\
        \textbf{\Name{}} & \cmark & \cmark & \cmark & \cmark & \cmark & \cmark & \cmark & \textbf{C/C++} \\
        \bottomrule
    \end{tabular}
    }
    \label{tab_libs}
\end{table*}

Existing libraries supporting learning or inference on PGMs include 
causal-learn~\citep{causal_learn}, 
Tetrad~\citep{tetrad},
ParallelPC~\citep{parapc},
bnlearn~\citep{bnlearn}, 
pcalg~\citep{pcalg},
FastInf~\citep{fastinf}, 
libDAI~\citep{libdai}, 
UnBBayes~\citep{unbbayes},
Factorie~\citep{factorie},
BNJ~\citep{bnj},
BNT~\citep{bnt},
SMILE\footnote{\url{https://www.bayesfusion.com/smile/}},   
and PNL\footnote{\url{https://sourceforge.net/projects/openpnl/}}. We provide a feature comparison of our \Name{} with these libraries in Table~\ref{tab_libs}. %In this table, we omit the libraries that have not been updated in the past two years in this table.

\section{Conclusion}

This paper introduces \textit{\Name{}}, a fast and parallel system designed for PGM learning and inference. \Name{} supports the fundamental PGM tasks of structure learning, parameter learning, exact inference and approximate inference with enhanced efficiency. Meanwhile, \Name{} provides modular building blocks, various functionalities, user-friendly interfaces, and comprehensive documentation to allow developers quick optimizations and extensions and ensure accessibility for non-experts.

% Acknowledgements and Disclosure of Funding should go at the end, before appendices and references

\acks{This research was funded by ARC Grant number DP190102443. Ajmal Mian is the recipient of an Australian Research Council Future Fellowship Award (project number FT210100268) funded by the Australian Government. Zeyi Wen is the corresponding author.}


\newpage

% \appendix
% \section{LALALA}

% \section{bbb}

\vskip 0.2in
\bibliography{main}

\begin{thebibliography}{47}
\providecommand{\natexlab}[1]{#1}
\providecommand{\url}[1]{\texttt{#1}}
\expandafter\ifx\csname urlstyle\endcsname\relax
  \providecommand{\doi}[1]{doi: #1}\else
  \providecommand{\doi}{doi: \begingroup \urlstyle{rm}\Url}\fi

\bibitem[Acid and de~Campos(2003)]{shd}
Silvia Acid and Luis~M de~Campos.
\newblock Searching for {B}ayesian network structures in the space of restricted acyclic partially directed graphs.
\newblock \emph{Journal of Artificial Intelligence Research}, 18:\penalty0 445--490, 2003.

\bibitem[Alfalahi et~al.(2023)Alfalahi, Dias, Khandoker, Chaudhuri, and Hadjileontiadis]{bnapp2}
Hessa Alfalahi, Sofia~B Dias, Ahsan~H Khandoker, Kallol~Ray Chaudhuri, and Leontios~J Hadjileontiadis.
\newblock A scoping review of neurodegenerative manifestations in explainable digital phenotyping.
\newblock \emph{npj Parkinson's Disease}, 9\penalty0 (1):\penalty0 49, 2023.

\bibitem[Amin et~al.(2021)Amin, Khan, Ahmed, and Imtiaz]{bnapp1}
Md~Tanjin Amin, Faisal Khan, Salim Ahmed, and Syed Imtiaz.
\newblock A data-driven {B}ayesian network learning method for process fault diagnosis.
\newblock \emph{Process Safety and Environmental Protection}, 150:\penalty0 110--122, 2021.

\bibitem[Bayoudh et~al.(2021)Bayoudh, Knani, Hamdaoui, and Mtibaa]{mnapp3}
Khaled Bayoudh, Raja Knani, Fay{\c{c}}al Hamdaoui, and Abdellatif Mtibaa.
\newblock A survey on deep multimodal learning for computer vision: advances, trends, applications, and datasets.
\newblock \emph{The Visual Computer}, pages 1--32, 2021.

\bibitem[Briganti et~al.(2022)Briganti, Scutari, and McNally]{bnapp3}
Giovanni Briganti, Marco Scutari, and Richard~J McNally.
\newblock A tutorial on {B}ayesian networks for psychopathology researchers.
\newblock \emph{Psychological methods}, 2022.

\bibitem[Carvalho et~al.(2010)Carvalho, Laskey, Costa, Ladeira, Santos, and Matsumoto]{unbbayes}
Rommel Carvalho, KB~Laskey, Paulo Costa, Marcelo Ladeira, La{\'e}cio Santos, and Shou Matsumoto.
\newblock Un{BB}ayes: modeling uncertainty for plausible reasoning in the semantic web.
\newblock \emph{Semantic Web}, pages 953--978, 2010.

\bibitem[Cheng and Druzdzel(2000)]{aisbn}
Jian Cheng and Marek~J Druzdzel.
\newblock {AIS-BN}: An adaptive importance sampling algorithm for evidential reasoning in large {B}ayesian networks.
\newblock \emph{Journal of Artificial Intelligence Research}, 13:\penalty0 155--188, 2000.

\bibitem[Chickering et~al.(2004)Chickering, Heckerman, and Meek]{chickering2004large}
Max Chickering, David Heckerman, and Chris Meek.
\newblock Large-sample learning of {B}ayesian networks is {NP}-hard.
\newblock \emph{Journal of Machine Learning Research}, 5, 2004.

\bibitem[Cooper(1990)]{bni}
Gregory~F Cooper.
\newblock The computational complexity of probabilistic inference using {B}ayesian belief networks.
\newblock \emph{Artificial intelligence}, 42\penalty0 (2-3):\penalty0 393--405, 1990.

\bibitem[Cousins et~al.(1993)Cousins, Chen, and Frisse]{sisv1}
Steve~B Cousins, William Chen, and Mark~E Frisse.
\newblock A tutorial introduction to stochastic simulation algorithms for belief networks.
\newblock \emph{Artificial Intelligence in Medicine}, 5\penalty0 (4):\penalty0 315--340, 1993.

\bibitem[Cui et~al.(2020)Cui, Lin, Pu, and Wang]{mnapp1}
Zhiyong Cui, Longfei Lin, Ziyuan Pu, and Yinhai Wang.
\newblock Graph {M}arkov network for traffic forecasting with missing data.
\newblock \emph{Transportation Research Part C: Emerging Technologies}, 117:\penalty0 102671, 2020.

\bibitem[Dagum and Luby(1993)]{Dagum1993}
Paul Dagum and Michael Luby.
\newblock Approximating probabilistic inference in {B}ayesian belief networks is {NP}-hard.
\newblock \emph{Artificial intelligence}, 60\penalty0 (1):\penalty0 141--153, 1993.

\bibitem[de~Waal and Joubert(2022)]{bnapp7}
Alta de~Waal and Johan~W Joubert.
\newblock Explainable {B}ayesian networks applied to transport vulnerability.
\newblock \emph{Expert Systems with Applications}, 209:\penalty0 118348, 2022.

\bibitem[Fung and Chang(1990)]{lw}
Robert Fung and Kuo-Chu Chang.
\newblock Weighing and integrating evidence for stochastic simulation in bayesian networks.
\newblock In \emph{Machine Intelligence and Pattern Recognition}, volume~10, pages 209--219. Elsevier, 1990.

\bibitem[Garzon et~al.(2023)Garzon, Ferreira, Z{\'o}zimo, Fortes, Ferreira, Pinheiro, and Reis]{bnapp5}
Juan~L Garzon, {\'O}scar Ferreira, AC~Z{\'o}zimo, CJEM Fortes, AM~Ferreira, LV~Pinheiro, and MT~Reis.
\newblock Development of a {B}ayesian networks-based early warning system for wave-induced flooding.
\newblock \emph{International Journal of Disaster Risk Reduction}, 96:\penalty0 103931, 2023.

\bibitem[Hacking et~al.(2006)]{prob}
Ian Hacking et~al.
\newblock \emph{The emergence of probability: A philosophical study of early ideas about probability, induction and statistical inference}.
\newblock Cambridge University Press, 2006.

\bibitem[Henrion(1988)]{pls}
Max Henrion.
\newblock Propagating uncertainty in {B}ayesian networks by probabilistic logic sampling.
\newblock In \emph{Machine Intelligence and Pattern Recognition}, volume~5, pages 149--163. Elsevier, 1988.

\bibitem[Hsu et~al.(2003)Hsu, Joehannes, Thornton, Perry, Haverkamp, Gettings, and Guo]{bnj}
William~H Hsu, R~Joehannes, JA~Thornton, Benjamin~B Perry, LM~Haverkamp, ND~Gettings, and H~Guo.
\newblock Bayesian network tools in {Java (BNJ)} v2. 0.
\newblock \emph{Kansas State University Laboratory for Knowledge Discovery in Databases}, 2003.

\bibitem[Jaimovich et~al.(2010)Jaimovich, Meshi, McGraw, and Elidan]{fastinf}
Ariel Jaimovich, Ofer Meshi, Ian McGraw, and Gal Elidan.
\newblock {FastInf}: An efficient approximate inference library.
\newblock \emph{Journal of Machine Learning Research}, 11:\penalty0 1733--1736, 2010.

\bibitem[Jiang et~al.(2022)Jiang, Wen, and Mian]{fast-bn}
Jiantong Jiang, Zeyi Wen, and Ajmal Mian.
\newblock Fast parallel {B}ayesian network structure learning.
\newblock In \emph{International Parallel and Distributed Processing Symposium}, pages 617--627. IEEE, 2022.

\bibitem[Jiang et~al.(2023)Jiang, Wen, Mansoor, and Mian]{fast-bni}
Jiantong Jiang, Zeyi Wen, Atif Mansoor, and Ajmal Mian.
\newblock Fast parallel exact inference on {B}ayesian networks.
\newblock In \emph{ACM SIGPLAN Annual Symposium on Principles and Practice of Parallel Programming}, pages 425--426, 2023.

\bibitem[Jiang et~al.(2024{\natexlab{a}})Jiang, Wen, Mansoor, and Mian]{fastbo}
Jiantong Jiang, Zeyi Wen, Atif Mansoor, and Ajmal Mian.
\newblock Efficient hyperparameter optimization with adaptive fidelity identification.
\newblock In \emph{Proceedings of the IEEE/CVF Conference on Computer Vision and Pattern Recognition}, 2024{\natexlab{a}}.

\bibitem[Jiang et~al.(2024{\natexlab{b}})Jiang, Wen, Mansoor, and Mian]{jiang2024fast}
Jiantong Jiang, Zeyi Wen, Atif Mansoor, and Ajmal Mian.
\newblock Fast inference for probabilistic graphical models.
\newblock In \emph{2024 USENIX Annual Technical Conference (USENIX ATC 24)}, 2024{\natexlab{b}}.

\bibitem[Kalisch et~al.(2012)Kalisch, M{\"a}chler, Colombo, Maathuis, and B{\"u}hlmann]{pcalg}
Markus Kalisch, Martin M{\"a}chler, Diego Colombo, Marloes~H Maathuis, and Peter B{\"u}hlmann.
\newblock Causal inference using graphical models with the {R} package pcalg.
\newblock \emph{Journal of Statistical Software}, 47\penalty0 (11):\penalty0 1--26, 2012.

\bibitem[Kokolakis and Nanopoulos(2001)]{h-distance}
George Kokolakis and Photis Nanopoulos.
\newblock Bayesian multivariate micro-aggregation under the {H}ellinger’s distance criterion.
\newblock \emph{Research in Official Statistics}, 4\penalty0 (1):\penalty0 117--126, 2001.

\bibitem[Koller and Friedman(2009)]{pgm}
Daphne Koller and Nir Friedman.
\newblock \emph{Probabilistic graphical models: principles and techniques}.
\newblock MIT press, 2009.

\bibitem[Lauritzen and Spiegelhalter(1988)]{jt}
Steffen~L Lauritzen and David~J Spiegelhalter.
\newblock Local computations with probabilities on graphical structures and their application to expert systems.
\newblock \emph{Journal of the Royal Statistical Society: Series B (Methodological)}, 50\penalty0 (2):\penalty0 157--194, 1988.

\bibitem[Le et~al.(2016)Le, Hoang, Li, Liu, Liu, and Hu]{parapc}
Thuc~Duy Le, Tao Hoang, Jiuyong Li, Lin Liu, Huawen Liu, and Shu Hu.
\newblock A fast {PC} algorithm for high dimensional causal discovery with multi-core pcs.
\newblock \emph{Transactions on Computational Biology and Bioinformatics}, 16\penalty0 (5):\penalty0 1483--1495, 2016.

\bibitem[Li et~al.(2023)Li, Ren, and Yang]{bnapp6}
Huanhuan Li, Xujie Ren, and Zaili Yang.
\newblock Data-driven {B}ayesian network for risk analysis of global maritime accidents.
\newblock \emph{Reliability Engineering \& System Safety}, 230:\penalty0 108938, 2023.

\bibitem[Lu et~al.(2021)Lu, Chen, Zhao, and Li]{mnapp2}
Yi~Lu, Yaran Chen, Dongbin Zhao, and Dong Li.
\newblock {MGRL}: Graph neural network based inference in a markov network with reinforcement learning for visual navigation.
\newblock \emph{Neurocomputing}, 421:\penalty0 140--150, 2021.

\bibitem[McCallum et~al.(2009)McCallum, Schultz, and Singh]{factorie}
Andrew McCallum, Karl Schultz, and Sameer Singh.
\newblock {FACTORIE}: Probabilistic programming via imperatively defined factor graphs.
\newblock \emph{Advances in Neural Information Processing Systems}, 22, 2009.

\bibitem[Mooij(2010)]{libdai}
Joris~M Mooij.
\newblock {libDAI}: A free and open source {C}++ library for discrete approximate inference in graphical models.
\newblock \emph{Journal of Machine Learning Research}, 11:\penalty0 2169--2173, 2010.

\bibitem[Murphy et~al.(2001)]{bnt}
Kevin Murphy et~al.
\newblock The {B}ayes net toolbox for {M}atlab.
\newblock \emph{Computing Science and Statistics}, 33\penalty0 (2):\penalty0 1024--1034, 2001.

\bibitem[Murphy et~al.(1999)Murphy, Weiss, and Jordan]{lbp}
Kevin~P. Murphy, Yair Weiss, and Michael~I. Jordan.
\newblock Loopy belief propagation for approximate inference: An empirical study.
\newblock In \emph{Conference in Uncertainty in Artificial Intelligence}, 1999.

\bibitem[Nayak et~al.(2022)Nayak, Kumar, Gupta, Suri, Naved, and Soni]{bnapp4}
Nihar~Ranjan Nayak, Sumit Kumar, Deepak Gupta, Ashish Suri, Mohd Naved, and Mukesh Soni.
\newblock Network mining techniques to analyze the risk of the occupational accident via {B}ayesian network.
\newblock \emph{International Journal of System Assurance Engineering and Management}, 13\penalty0 (1):\penalty0 633--641, 2022.

\bibitem[Ramsey et~al.(2018)Ramsey, Zhang, Glymour, Romero, Huang, Ebert-Uphoff, Samarasinghe, Barnes, and Glymour]{tetrad}
Joseph~D Ramsey, Kun Zhang, Madelyn Glymour, Ruben~Sanchez Romero, Biwei Huang, Imme Ebert-Uphoff, Savini Samarasinghe, Elizabeth~A Barnes, and Clark Glymour.
\newblock {TETRAD}—a toolbox for causal discovery.
\newblock In \emph{International Workshop on Climate Informatics}, 2018.

\bibitem[Salinas et~al.(2022)Salinas, Seeger, Klein, Perrone, Wistuba, and Archambeau]{synetune}
David Salinas, Matthias Seeger, Aaron Klein, Valerio Perrone, Martin Wistuba, and Cedric Archambeau.
\newblock Syne tune: A library for large scale hyperparameter tuning and reproducible research.
\newblock In \emph{International Conference on Automated Machine Learning}, pages 16--1. PMLR, 2022.

\bibitem[Scutari(2014)]{bnlearn}
Marco Scutari.
\newblock {B}ayesian network constraint-based structure learning algorithms: Parallel and optimised implementations in the bnlearn {R} package.
\newblock \emph{arXiv preprint arXiv:1406.7648}, 2014.

\bibitem[Shachter and Peot(1990)]{sis}
Ross~D Shachter and Mark~A Peot.
\newblock Simulation approaches to general probabilistic inference on belief networks.
\newblock In \emph{Machine Intelligence and Pattern Recognition}, volume~10, pages 221--231. Elsevier, 1990.

\bibitem[Spirtes and Glymour(1991)]{pc}
Peter Spirtes and Clark Glymour.
\newblock An algorithm for fast recovery of sparse causal graphs.
\newblock \emph{Social Science Computer Review}, 9\penalty0 (1):\penalty0 62--72, 1991.

\bibitem[Spirtes et~al.(2000)Spirtes, Glymour, Scheines, and Heckerman]{pc2}
Peter Spirtes, Clark~N Glymour, Richard Scheines, and David Heckerman.
\newblock \emph{Causation, prediction, and search}.
\newblock MIT press, 2000.

\bibitem[Wang et~al.(2023)Wang, Xu, Sun, and Liu]{mnapp5}
Chuanyuan Wang, Shiyu Xu, Duanchen Sun, and Zhi-Ping Liu.
\newblock {ActivePPI}: quantifying protein--protein interaction network activity with {M}arkov random fields.
\newblock \emph{Bioinformatics}, 39\penalty0 (9):\penalty0 btad567, 2023.

\bibitem[Xu et~al.(2024)Xu, Wang, and Zhang]{mnapp4}
Yangyang Xu, Zengmao Wang, and Xiaoping Zhang.
\newblock Leveraging spatial residual attention and temporal {M}arkov networks for video action understanding.
\newblock \emph{Neural Networks}, 169:\penalty0 378--387, 2024.

\bibitem[Yuan and Druzdzel(2003)]{episbn}
Changhe Yuan and Marek~J. Druzdzel.
\newblock An importance sampling algorithm based on evidence pre-propagation.
\newblock In \emph{Conference in Uncertainty in Artificial Intelligence}, pages 624--631, 2003.

\bibitem[Yuan and Druzdzel(2006)]{episbn2}
Changhe Yuan and Marek~J. Druzdzel.
\newblock Importance sampling algorithms for {B}ayesian networks: Principles and performance.
\newblock \emph{Mathematical and Computer Modelling}, 43\penalty0 (9-10):\penalty0 1189--1207, 2006.

\bibitem[Zhang and Poole(1994)]{Zhang1994}
Nevin~L Zhang and David Poole.
\newblock A simple approach to {B}ayesian network computations.
\newblock In \emph{Canadian Conference on Artificial Intelligence}, 1994.

\bibitem[Zheng et~al.(2024)Zheng, Huang, Chen, Ramsey, Gong, Cai, Shimizu, Spirtes, and Zhang]{causal_learn}
Yujia Zheng, Biwei Huang, Wei Chen, Joseph Ramsey, Mingming Gong, Ruichu Cai, Shohei Shimizu, Peter Spirtes, and Kun Zhang.
\newblock Causal-learn: Causal discovery in python.
\newblock \emph{Journal of Machine Learning Research}, 25\penalty0 (60):\penalty0 1--8, 2024.

\end{thebibliography}

\end{document}